%% file: main.tex
\documentclass[10pt,twocolumn,letterpaper]{article}

\usepackage{cvpr}

\usepackage{graphicx}
\usepackage{amsmath}
\usepackage{amssymb}
\usepackage{booktabs}
\usepackage[euler]{textgreek}

\usepackage{graphbox}
\usepackage{paralist}

\usepackage{makecell}
\usepackage{tabularx}
\usepackage{multirow}
\usepackage{multicol}

\usepackage[T1]{fontenc}    
\usepackage{amsfonts}       
\usepackage{pifont}
\usepackage{dsfont}

\usepackage{listings}
\usepackage{setspace}
\usepackage{color}
\usepackage[dvipsnames,svgnames]{xcolor}
\usepackage{colortbl}
\definecolor{Gray}{gray}{0.85}
\definecolor{LightCyan}{rgb}{0.88,1,1}

\usepackage[utf8]{inputenc} 
\usepackage{url}            
\usepackage{mathtools}
\usepackage{nicefrac}       
\usepackage{ragged2e}
\usepackage{mwe}

\usepackage[normalem]{ulem}

\usepackage{url}
\usepackage{color}
\usepackage[pagebackref,breaklinks,colorlinks,citecolor=MyGreen]{hyperref}
\usepackage{algorithm}
\usepackage{algpseudocode}

\definecolor{darkergreen}{RGB}{21, 152, 56}
\definecolor{red2}{RGB}{252, 54, 65}
\usepackage[capitalize]{cleveref}
\crefname{section}{Sec.}{Secs.}
\Crefname{section}{Section}{Sections}
\Crefname{table}{Table}{Tables}
\crefname{table}{Tab.}{Tabs.}

\usepackage[accsupp]{axessibility}  

\def\cvprPaperID{6847} 
\def\confName{CVPR}
\def\confYear{2024}

\begin{document}

\include{macro}

\title{Towards Scalable 3D Anomaly Detection and Localization: 
A Benchmark via 3D Anomaly Synthesis and A Self-Supervised Learning Network }
\author{$\text{Wenqiao Li}^{1}$\quad $\text{Xiaohao Xu}^{2}$
\quad $\text{Yao Gu}^{1}$\quad $\text{Bozhong Zheng}^{1}$\quad $\text{Shenghua Gao}^{1\ast}$\quad $\text{Yingna Wu}^{1\ast}$\\
$^{1}$ShanghaiTech University \quad $^{2}$University of Michigan, Ann Arbor\\
{\tt\small \{liwq2022\}@shanghaitech.edu.cn}
}
\maketitle

\input{latex/sections/0_abstract}
 \input{latex/sections/1_introduction}
 \input{latex/sections/2_related_works}

\input{latex/sections/3_proposed_methods}
 \input{latex/sections/4_proposed_methods}
 \input{latex/sections/5_experiments}

 \input{latex/sections/6_conclusion}

{\small
\bibliographystyle{ieee_fullname}
\bibliography{egbib}
}

\end{document}

%% file: macro.tex

\long\def\ignorethis#1{}



\newcommand{\img}[1]{\mathbf{I}_{\text{#1}}}
\newcommand{\paren}[1]{\left( #1 \right)}
\newcommand{\bparen}[1]{\left[ #1 \right]}
\newcommand{\feature}[1]{\phi \paren{#1}}
\newcommand{\normtwo}[1]{\lVert #1 \rVert_2^2}
\newcommand{\normone}[1]{\left\lVert #1 \right\rVert_1}

\newcommand{\Paragraph}[1]{\vspace{1mm}\noindent\textbf{#1}}

\newcommand{\figref}[1]{Figure~\ref{fig:#1}}
\newcommand{\tabref}[1]{Table~\ref{tab:#1}} 
\newcommand{\itmref}[1]{[\ref{itm:#1}]}     
\newcommand{\eqnref}[1]{\eqref{eq:#1}}
\newcommand{\secref}[1]{Section~\ref{sec:#1}}
\newcommand{\eqmain}[1]{(\textcolor{blue}{#1})}
\newcommand{\fakeref}[1]{\textcolor{MyGreen}{#1}}
\newcommand{\fakeeqref}[1]{\textcolor{MyGreen}{(#1)}}

\newcommand{\mb}[1]{\mathbf{#1}}
\newcommand{\bs}[1]{\boldsymbol{#1}}
\newcommand{\n}{\mbox{\qquad}}              
\newcommand{\red}[1]{{\color{red}#1}}

\newcommand{\ignore}[1]{}   
\newcommand{\cmt}[1]{\begin{sloppypar}\large\textcolor{red}{#1}\end{sloppypar}}

\newcommand{\TODO}[1]{\textcolor{red}{[TODO]\{#1\}}}
\newcommand{\todo}[1]{\textcolor{red}{#1}}
\newcommand{\torevise}[1]{\textcolor{blue}{#1}}
\newcommand{\revise}[1]{\textcolor{blue}{#1}}
\newcommand{\copied}[1]{\textcolor{red}{[COPIED: #1]}}

\newcommand{\needref}{[\textcolor{blue}{put}, \textcolor{blue}{some}, \textcolor{blue}{references}]}

\newcommand{\tabspace}{\vspace{-2mm}}
\newcommand{\tabxspace}{\vspace{-4mm}}
\newcommand{\figspace}{\vspace{-2mm}}
\newcommand{\figxspace}{\vspace{-3mm}}
\newcommand{\best}[1]{{\textcolor{red}{\textbf{#1}}}}
\newcommand{\secondbest}[1]{{\textcolor{blue}{\underline{#1}}}}

\newcommand{\ts}{\textsuperscript}

\newcommand{\set}[1]{\{#1\}}


\def\PE{\Phi}

\newcommand{\mpage}[2]
{
\begin{minipage}{#1\linewidth}\centering
#2
\end{minipage}
}

\newcommand{\replace}[2]{\textcolor{red}{\sout{#1}} \textcolor{blue}{#2}}
\newcommand{\fix}[1]{\textcolor{red}{#1}}
\newcommand{\proposed}{UDMM}

\newcommand{\bsd}{MSD}

\newcommand{\blurhigh}{\mathbf{I}_{\hat{\text{B}}}}
\newcommand{\blurlow}{\mathbf{I}_{\hat{\text{B}},\hat{\text{LR}}}}

\definecolor{MyGreen}{cmyk}{100, 0, 100, 0}

\newcommand{\citenumber}[1]{[\textcolor{MyGreen}{#1}]}
\newcommand{\refnumber}[1]{\textcolor{red}{#1}}

\newcolumntype{L}[1]{>{\raggedright\let\newline\\\arraybackslash\hspace{0pt}}m{#1}}
\newcolumntype{C}[1]{>{\centering\let\newline\\\arraybackslash\hspace{0pt}}m{#1}}
\newcolumntype{R}[1]{>{\raggedleft\let\newline\\\arraybackslash\hspace{0pt}}m{#1}}

\newcommand{\jaeha}[1]{{\textcolor{BurntOrange}{\textbf{Jaeha: }#1}}}
\newcommand{\joonkyu}[1]{{\textcolor{blue}{\textbf{Joonkyu: }#1}}}
\newcommand{\younguk}[1]{{\textcolor{ProcessBlue}{\textbf{Younguk: }#1}}}


%% file: latex/sections/0_abstract.tex
\begin{abstract}
Recently, 3D anomaly detection, a crucial problem involving fine-grained geometry discrimination,  is getting more attention. However, the lack of abundant real 3D anomaly data limits the scalability of current models. 
To enable scalable anomaly data collection, we propose a 3D anomaly synthesis pipeline to adapt existing large-scale 3D models for 3D anomaly detection. Specifically, we construct a synthetic dataset, \textit{i.e.}, \textbf{Anomaly-ShapeNet}, based on ShapeNet. Anomaly-ShapeNet consists of 1600 point cloud samples under 40 categories, which provides a rich and varied collection of data, enabling efficient training and enhancing adaptability to industrial scenarios.
Meanwhile, to enable scalable representation learning for 3D anomaly localization, we propose a self-supervised method, \textit{i.e.}, Iterative Mask Reconstruction Network (\textbf{IMRNet}). During training, we propose a geometry-aware sample module to preserve potentially anomalous local regions during point cloud down-sampling. Then, we randomly mask out point patches and sent the visible patches to a transformer for reconstruction-based self-supervision. During testing, the point cloud repeatedly goes through the Mask Reconstruction Network, with each iteration's output becoming the next input. By merging and contrasting the final reconstructed point cloud with the initial input, our method successfully locates anomalies. Experiments show that IMRNet outperforms previous state-of-the-art methods, achieving 66.1\% in I-AUC on Anomaly-ShapeNet dataset and 72.5\% in I-AUC on Real3D-AD dataset. Our dataset will be released at https://github.com/Chopper-233/Anomaly-ShapeNet.

\end{abstract}

%% file: latex/sections/1_introduction.tex
\section{Introduction}

\begin{figure}[t]
  \centering
  \includegraphics[width=1.03\linewidth]{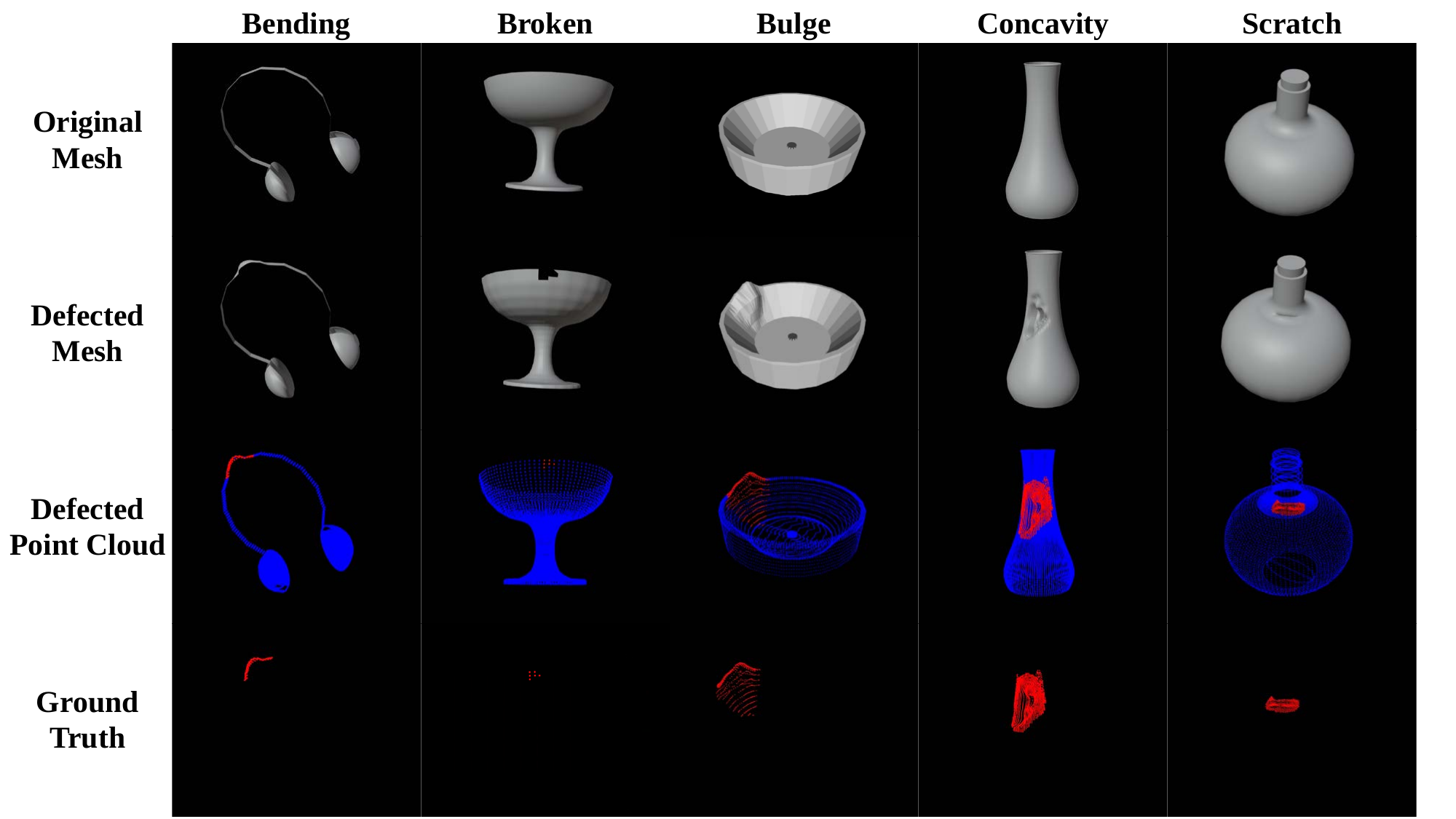}
       \caption{\textbf{Examples of the proposed Anomaly-ShapeNet}. The first and second rows are the original mesh and the subdivided mesh. The third row is synthetic defected point cloud. The fourth row is the Ground-Truth of the anomalous region.}
   \vspace{-10pt}
   \label{fig:examples}
\end{figure}

The 3D anomaly detection task is crucial in object-centered industrial quality inspection, where high accuracy and detection rates are often required. The goal is to identify anomaly regions and locate abnormal point clouds within the 3D context. Image-based anomaly detection algorithms under fixed perspectives~\cite{bergmann2020uninformed,cohen_2020_spade,Defard_2021_PaDiM,Gudovskiy_2022_CFLOW,Mishra_2020_piade,Reiss_2021_PANDA,Rippel_2021_Gaussian,roth2022towards,defard2021padim,zavrtanik2021draem,li2021cutpaste,Ristea-CVPR-2022,zou2022spot,roth2022towards,cao2023collaborative} have limitations due to blind spots and do not perform as desired in object-centered scenarios. Consequently, researchers have increasingly focused on 3D information for anomaly detection~\cite{Bergmann_2022_mvtec_3dad}.

With advancements in 3D sensor technology, several datasets such as MvtecAD-3D~\cite{Bergmann_2022_mvtec_3dad}, Real3D-AD~\cite{liu2023real3d}, and MAD~\cite{zhou2023pad} have been created to meet the increasing demand for 3D anomaly detection. MvtecAD-3D is designed for anomaly detection in scenarios with a single-angle camera, while the MAD dataset focuses on multi-pose anomaly detection. Only Real3D-AD specifically addresses anomaly detection on complete point clouds. Since 3D point clouds obtained from a 3D scanner generally contain more morphological information compared to data from multiple cameras, our primary objective is to advance the field of point cloud-based anomaly detection tasks.

In the field of point cloud anomaly detection, there are two main issues that need to be addressed: the lack of diverse distribution datasets and the need for more effective deep learning-based approaches. Firstly, the current high-quality real-world 3D point anomaly detection dataset, Real3D-AD, has limitations in terms of the variety of normal and abnormal samples and the excessive variation in point sizes. These factors make it challenging to apply reconstruction-based~\cite{Bergmann_2019_SSIM_AE} or deep learning feature extraction-based~\cite{bergmann2020uninformed} algorithms widely. Secondly, the prevailing approaches for point cloud anomaly detection heavily rely on traditional feature processing operators and models that are pretrained to extract features or transform data to 2D for processing~\cite{bergmann2020uninformed,cao2023complementary}. This approach fails to fully exploit the potential of point cloud data and leads to significant feature domain misalignment.

To address the current dataset limitations, we have created the Anomaly-ShapeNet dataset by synthesizing data based on ShapeNet~\cite{chang2015shapenet}, which is a widely used dataset for point cloud processing. The Anomaly-ShapeNet dataset consists of 1600 samples belonging to 40 classes. It includes six typical anomaly types: bulge, concavity, hole, break, bending, and crack. The number of points in each sample ranges from 8000 to 30,000. The anomalous portion constitutes between 1\% and 10\% of the entire point cloud. Importantly, our synthetic anomaly dataset contains realistic and diverse examples of anomalies. Figure~\ref{fig:examples} showcases some examples from the Anomaly-ShapeNet dataset.

To address the challenges in 3D point cloud anomaly detection, we propose a self-supervised method called IMRNet, which is based on iterative mask reconstruction. IMRNet comprises three main components: (1) the Geometry-Aware Sample module (GPS), (2) the Point-Patch Mask Reconstruction network (PMR), and (3) the Dense Feature Concatenation and Comparison module (DFC).
In the IMRNet framework, before passing the anomalous point clouds through the reconstruction network, we employ the GPS module to extract keypoints based on the geometry structure of the point clouds. This ensures that we sample the anomaly points as comprehensively as possible. These sampled keypoints are then transformed into point-patch groups using the K-nearest neighbor (KNN) operation and sent to the PMR network in an iterative manner. The PMR network performs the mask reconstruction process on these point-patch groups. 
After the PMR network, both the reconstructed normal sample and the input point cloud are sent to the DFC module. The DFC module compares the features and points of both the reconstructed normal sample and the input point cloud to obtain the anomaly score. This comparison is performed in both the feature space and the point space, enabling a comprehensive evaluation of the anomaly.
Overall, the IMRNet method combines the GPS module, the PMR network, and the DFC module to achieve effective self-supervised 3D point cloud anomaly detection through iterative mask reconstruction and feature-based comparison.

By combining the Anomaly-ShapeNet dataset and the IMRNet method, our work aims to contribute to the field of scalable 3D anomaly detection. We summarize our contributions as follows:
\begin{itemize}
    \item We present a diverse and high-quality synthetic 3D anomaly dataset, Anomaly-ShapeNet, with accurate 3D annotations. 
    \item We propose the IMRNet, a novel 3D point-cloud anomaly detection method, with an Geometry-aware Point-cloud Sample module, an iterative Point-patch Mask Reconstruction network and a dense feature comparision module, which outperforms the SoTA methods on Anomaly-ShapeNet and Real3D-AD.
    \item We demonstrate that the proposed Geometry-aware Point-cloud Sample can help extract important anomaly points in 3D point clouds more effectively  and the proposed PMR network can learn better representation of the 3D anomaly datasets.
\end{itemize}

%% file: latex/sections/2_related_works.tex
\input{latex/tables/5.3/Dataset_comparision}

\section{Related works}
\noindent\textbf{2D Anomaly Detection.}
Recent advances in unsupervised anomaly detection for two-dimensional data, such as RGB or grayscale images, have shown significant progress~\cite{pang2021adreview,ehret_review_paper_2019,xie_im_iad_2023}. Traditional approaches often rely on autoencoders~\cite{Bergmann_2019_SSIM_AE,Hong_2020_DecentralizationLoss,Wang_2020_LatentSpaceResampling} and generative adversarial networks (GANs)~\cite{Carrara_2021_CBiGAN,Potter_2020_pandanet,Schlegl_2019_fAnoGan}, typically employing random weight initialization. Alternatively, methods leveraging pretrained network descriptors for anomaly detection often surpass those trained from scratch~\cite{bergmann2020uninformed,cohen_2020_spade,Defard_2021_PaDiM,Gudovskiy_2022_CFLOW,Mishra_2020_piade,Reiss_2021_PANDA,Rippel_2021_Gaussian,roth2022towards,defard2021padim,zavrtanik2021draem,li2021cutpaste,Ristea-CVPR-2022,zou2022spot,roth2022towards,cao2023collaborative,xie2023pushing,rudolph2023asymmetric,cao2022semi}. Noteworthy examples include Self-Taught anomaly detection (ST)~\cite{bergmann2020uninformed}, which aligns features from a pre-trained teacher network with a randomly initialized student network to identify anomalies, PatchCore~\cite{roth2022towards} that utilizes a memory-bank approach for modeling normal data distributions, and Cutpaste~\cite{li2021cutpaste}, introducing a novel data augmentation strategy for self-supervised anomaly detection.
The emergence of large-scale foundation models~\cite{radford2021learning,kirillov2023segment} has spurred new methodologies that capitalize on the robust zero-shot generalization capabilities of these models for anomaly detection tasks~\cite{cao2023segment,jeong2023winclip,chen_zero_few_shot_2023}. In our work, we extend the principles of reconstruction-based self-supervised learning from 2D to 3D contexts. We introduce a novel self-supervised network employing a masked reconstruction mechanism to advance scalable 3D anomaly detection.

\noindent\textbf{3D Anomaly Detection.} The domain of 3D anomaly detection remains less advanced compared to its 2D counterpart, hindered by intrinsic challenges such as data sparsity, increased dimensionality, and prevalent noise. Bergmann \textit{et al.}~\cite{bergmann2023anomaly} introduced a seminal approach wherein a teacher-student network paradigm is employed during training to align point-cloud features, which are then compared in testing to identify anomalies. Horwitz \textit{et al.}~\cite{horwitz2023back} leveraged classical 3D descriptors coupled with K-Nearest Neighbors (KNN) for anomaly detection. Although AST~\cite{rudolph2023asymmetric} demonstrated efficacy in certain scenarios, its primary focus on background suppression via depth information led to the omission of finer detail anomalies. The M3DM framework~\cite{wang2023multimodal} innovatively combines 3D point data with conventional imaging for enhanced decision-making. CPMF~\cite{cao2023complementary} introduced a novel methodology that integrates a memory bank approach with KNN and enriches the detection process by rendering 3D data into multi-view 2D images. Conversely, EasyNet~\cite{chen2023easynet} presents a straightforward mechanism for 3D anomaly detection, circumventing the need for pre-training.
Nonetheless, the scarcity of robust 3D anomaly detection datasets~\cite{liu2023real3d,Bergmann_2022_mvtec_3dad,zhou2023pad} constrains the scalability and generalizability of these models across varied 3D anomaly detection contexts. In light of this, our work aims to devise a 3D anomaly synthesis pipeline, enhancing the volume and diversity of data necessary for the development of more generalized 3D anomaly detection models.

%% file: latex/tables/5.3/Dataset_comparision.tex
\begin{table*}[t]
\centering
\scriptsize

\renewcommand\arraystretch{0.8}
\resizebox{0.95\linewidth}{!}{
\begin{tabular}{@{}cccc|ccccc@{}}
\toprule
\multirow{1}{*}{Datasets} &
  \multirow{1}{*}{Year} &
  \multirow{1}{*}{Type} &
  \multirow{1}{*}{Modality}  & $\#$Class   & $\#$Anomaly Types   & Number & Point Range    \\ 
    \midrule

MVTecAD-3D\cite{Bergmann_2022_mvtec_3dad}          & 2021 & Real     & RGB/D    & 10      & 3-5    & 3604      &10$K$-30$K$   \\
Eyecandies\cite{bonfiglioli2022eyecandies}          & 2022 & Syn     & RGB/D/N & 10     & 3   &15500    &-   \\
MAD\cite{zhou2023pad}                & 2023 & Syn+Real     & RGB     & 20     &3     & 10133    &-   \\ \midrule
Real3D-AD\cite{liu2023real3d}           & 2023 & Real     & Point Cloud  & 12     & 2        & 1200   & 35$K$-780$K$  \\
\textbf{Anomaly-ShapeNet (Ours)}        & 2023 & Syn & Point Cloud     & 40     & 6    & 1600    & 8$K$-30$K$   \\ \bottomrule
\end{tabular}}
\caption{\textbf{Comparison between the proposed Anomaly-ShapeNet and existing mainstream 3D anomaly detection datasets.} }
\label{tab:1}
\end{table*}

%% file: latex/sections/3_proposed_methods.tex
\section{Anomaly-ShapeNet 3D Dataset Synthesis}
\label{sec:Anomaly-ShapeNet dataset}

%
%
%
%

%

\noindent\textbf{Pipeline Overview.} As is shown in Figure \ref{fig:manufacture}, the overall pipeline for constructing our Anomaly-ShapeNet dataset  consists of three components: mesh subdivision, defects carving, and ground truth generation.

\noindent\textbf{Normal Data Sampling.}  To generate our dataset's original normal samples, we utilized the ShapeNet dataset~\cite{chang2015shapenet}, which is renowned for its diversity and high quality and is commonly employed in tasks such as point cloud segmentation, completion, and classification.  For our data source, we specifically selected normal samples from the sub-dataset ShapeNetcoreV2 of the ShapeNet.

\noindent\textbf{Anomaly Data Synthesis.}  We developed a point cloud refinement module to enhance the limited number of points and faces found in certain point clouds from the ShapeNet dataset. To introduce more realistic defects, we utilized \textit{Blender}, a widely-used software in the industrial design domain to sculpt various defects. Blender is an open-source industrial design software offering extensive features such as sculpting, refining, cropping, and various editing modes, which contributes to its popularity in this field. After acquiring the abnormal samples, we employed \textit{CloudCompare}, point editing software, to obtain the ground truths.

\noindent\textbf{Dataset Statistics.} Anomaly-ShapeNet comprises training folders with four normal samples per class and testing folders with 28 to 40 samples, including both normal and abnormal instances. Sample point clouds range from 8,000 to 30,000 points, compatible with prevalent point cloud models and minimizing the need for downsampling. This dataset provides a broader category variety and a more application-friendly number of points compared to Real3D-AD~\cite{wang2023multimodal}, the sole 3D point cloud anomaly detection dataset from 3D scans. It expands the scope of Real3D-AD, offering increased diversity and a wider range of anomaly types. It aligns with the data volume preferred by previous algorithms, thereby facilitating advancements in anomaly detection and reducing computational overhead. The comparison with leading 3D anomaly datasets is detailed in Table~\ref{tab:1}.


\input{latex/figures/manufacture/manufacture}
\input{latex/sections/figures/model}

%% file: latex/figures/manufacture/manufacture.tex
\begin{figure}[t]
\begin{center}
\subfloat{\includegraphics[width=\linewidth]{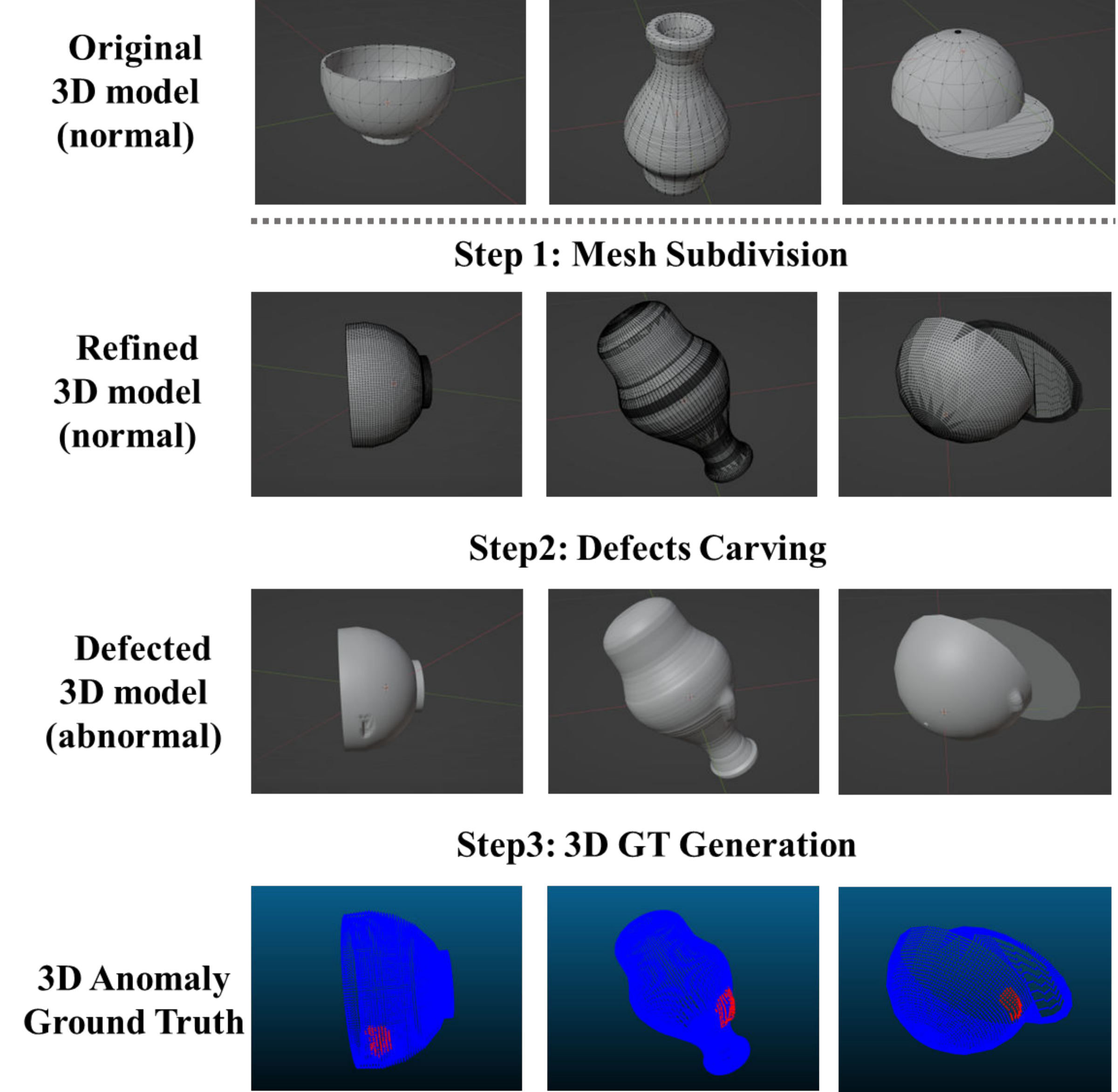}}\end{center}
\vspace{-3mm}
\caption{\textbf{Pipeline for anomaly synthesis built upon the Anomaly-ShapeNet Dataset.}
Selected normal samples are processed through a mesh subdivision module to attain a more uniform point cloud distribution. We employ \textit{Blender} to introduce defects into the refined samples, and utilize \textit{CloudCompare} software to acquire 3D anomaly ground truth.}

%

\vspace{-2mm}
\label{fig:manufacture}
\end{figure}

%% file: latex/sections/figures/model.tex
\begin{figure*}[t]
\begin{center}
\includegraphics[width=0.94\linewidth]{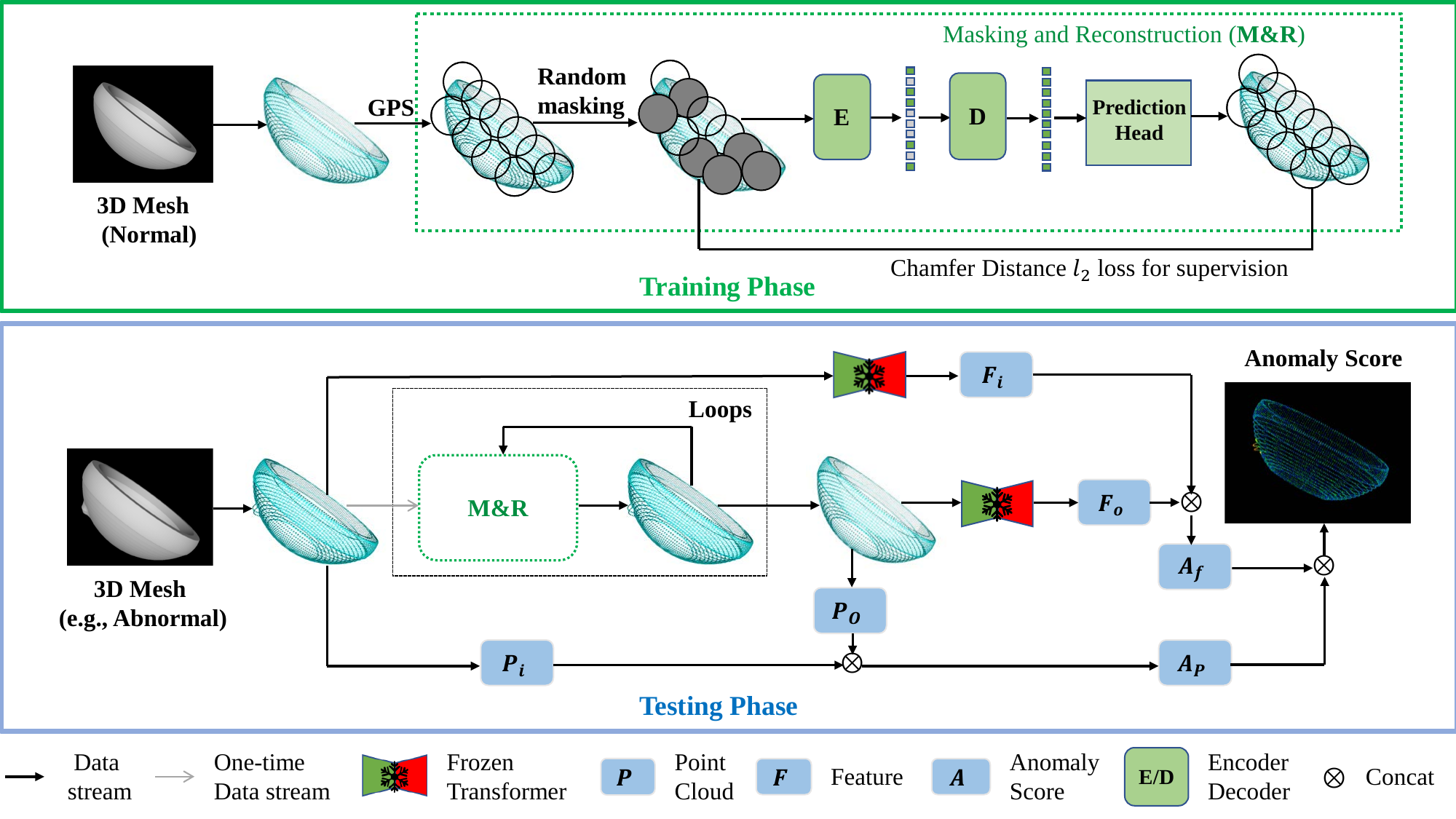}
\end{center}
\vspace{-7mm}
\caption{\textbf{Overview of the Iterative Mask Reconstruction Network (IMRNet) Pipeline.} (a) \textit{Training Phase:} The standard training point cloud is initially converted to point-patch format using the Geometry-aware Point-cloud Sampling (GPS) module. Following this, random masking is applied to the point-patches, which are then reconstructed by a network comprising an Autoencoder-based Transformer and a lightweight prediction head, operating in a self-supervised paradigm. (b) \textit{Testing Phase:} The input point cloud is subjected to a reconstruction process mirroring the training procedure. The reconstructed point cloud is cyclically fed back into the reconstruction network as input for several iterations. Ultimately, a comparative analysis is performed between the reconstructed and the original point clouds at both the point cloud and feature levels to derive the final anomaly score map.}

\vspace{-3mm}
\label{fig:model}
\end{figure*}

%% file: latex/sections/4_proposed_methods.tex
\section{Self-Supervised  Representation Learning for 3D Anomaly Detection: IMRNet}
Figure~\ref{fig:model} shows the overall architecture of our IMRNet.
Our IMRNet, which consists of three modules: GPS, PMR, and DFC, successfully detect and locate the anomaly in the abnormal samples. The details of each module will be illustrated in the following sections.

\subsection{Geometry-aware Point Cloud Sampling}

In point cloud processing, uniform sampling or farthest point sampling is commonly used. However, when it comes to point cloud anomaly detection, arbitrary sampling methods can lead to ambiguous representations of the anomaly structures. In our geometry aware sampling module, we address this issue by first calculating the geometry features of the points to adaptively sample the point cloud.

\noindent\textbf{Geometry feature extraction.} Given a point cloud ($P$) as input, we define the set of neighboring points of a certain point $P_i$ within a radius $r$ as $\mathcal{N}_i$ (where $|\mathcal{N}_i| = N$). To compute the normal vector $\mathbf{N}_i$ for each point ($P_i$), we employ a local surface fitting method. The normal vector $\mathbf{N}_i$ can be obtained by solving the following equation:
\begin{equation}
\min_{\mathcal{N}_i} \sum_{P_j \in \mathcal{N}_i} \left| \mathbf{N}_i \cdot \mathbf{v}_j - d \right|^2
\end{equation}
where ($\mathbf{v}_j$) is the vector from $P_i$ to $P_j$, $\mathbf{N}_i \cdot \mathbf{v}_j$ denotes the dot product between the normal vector and the vector $\mathbf{v}_j$, and $d$ represents the signed distance from the plane defined by the normal vector $\mathbf{N}_i$ to the point $P_j$.

The curvature of a point can be calculated using the normal vector and the eigenvalues of the covariance matrix $\mathbf{C}_i$ of its neighboring points $\mathcal{N}_i$). The curvature value ($K_i$) is given by:
\begin{equation}
K_i = \frac{{\min(\lambda_1, \lambda_2)}}{{\lambda_1 + \lambda_2 + \epsilon}}
\end{equation}
where ($\lambda_1$) and ($\lambda_2$) are the eigenvalues of $\mathbf{C}_i$ and $\epsilon$ is a small positive constant to avoid division by zero.

To quantify the rate of change in normal vectors and curvature between two points ($P_i$ and $P_j$), we can define the following formulas:
\begin{equation}
R_{\text{norm}}(P_i, P_j) = \frac{{\left| \mathbf{N}_i - \mathbf{N}j \right|}}{{\left| \mathbf{v}_{ij} \right|}}
\end{equation}

\begin{equation}
R_{\text{curv}}(P_i, P_j) = \left| K_i - K_j \right|
\end{equation}
where $\mathbf{v}_{ij}$ is the vector from $P_i$ to $P_j$, $\mathbf{N}_i$ and $\mathbf{N}_j$ are the normal vectors at each point, and $K_i$ and $K_j$ are the curvature values at each point.

\noindent\textbf{Geometry-aware sampling.} During the sampling process of the input point cloud, we employ a geometry-aware point cloud sampling approach. To achieve this, we introduce a rate of change memory bank $\mathcal{M}$ that captures the local and global rates of change in the point cloud. For each point $P_i$, the memory bank stores the rate of change value $R_i$ computed based on the differences in normal vectors and curvature values between neighboring points. Given the rate of change in normal vectors and curvature ($R_{\text{norm}}$ and $R_{\text{curv}}$), the overall rate of change value $R$ is computed as the average of the rate of change values with its neighboring points:
\begin{equation}
    R_i = \frac{1}{|\mathcal{N}_i|} \sum_{P_j \in N_i} \left( R_{\text{norm}}(P_i, P_j) + R_{\text{curv}}(P_i, P_j) \right)
\end{equation}

To prioritize high-rate-of-change points, we sort them based on values in the memory bank $\mathcal{M}$. Lower ranks indicate higher rates of change, capturing significant anomalous structures. We select points with greater rates of change using a threshold $\tau$. By sampling points with ranks from 1 to $\lfloor \tau \cdot N \rfloor$, where $N$ is the total point count, we capture more points with significant rates of change. The sampling process to derive the final sampled point set ($\mathcal{S}$) can be represented by the following equation:
\begin{equation}
\mathcal{S} = \{P_k| \text{Rank}(P_k) \leq \lfloor \tau \cdot N \rfloor\}
\end{equation}

By incorporating this geometry-aware point cloud sampling strategy, we enhance the accuracy and effectiveness of point cloud anomaly detection by ensuring that the sampled points represent the underlying anomaly structures with their distinctive geometric characteristics.

\subsection{Iterative Mask Reconstruction}

In 2D anomaly detection field, splitting normal images into patches of the same size and randomly masking them before reconstructing the masked patches back to normal is common. The underlying idea is the reconstruction network only receives the normal patches and learns only the normal feature distribution during training. During testing, input anomalous images are masked and can be restored to normal images. Different from images, point cloud can not be divided into regular patches like images. Furthermore, due to the unordered nature of point clouds, the reconstruction error cannot be directly computed using mean squared error (MSE) or structural similarity (SSIM). Based on these properties, we propose the PMR module, which mainly consists of three components: point-patch generation, random masking and embedding, and reconstruction target.

\noindent\textbf{Point-patch generation.}
Following Point-MAE~\cite{pang2022masked}, the input point cloud is divided into irregular overlapping patches via our Geometry-aware Point Sampling (GPS) and K-Nearest Neighborhood (KNN).  Formally, given an point cloud $P_i$ with $N$ points, ${P_i} \in \mathbb{R}^{N \times 3}$, GPS is applied to sample $C$ points as the patch centers.  Based on $C$, KNN selects $k$ nearest points around each center and formulate the point patches $P$. $C$ and $P$ are formulated as follows:
\begin{equation}
    C = GPS(P_i),  {C} \in\mathbb{R}^{n \times 3}
\end{equation}
\begin{equation}
    P = KNN(P_i,C),  {P} \in\mathbb{R}^{n \times k  \times  3}
\end{equation}

It should be noted that, in our point patches, each center point represents its neighborhood like 2D patchcore~\cite{roth2022towards}.This not only leads to better convergence but also facilitates the detection and localization of anomalies.

\noindent\textbf{Random masking and embedding}.
With a masking ratio $m$ at 40\%, we randomly select a set of masked patches $M \in \mathbb{R}^{mn \times k \times 3}$, which is also used as the ground truth for reconstruction. After ramdomly masking, the visible points could be illustrated as:
\begin{equation}
    P_{vis} = P \odot (1-M)
\end{equation}
where $\odot$ denotes the spatial element-wise product.

To transform the visible patches and masked patches to tokens, we implement the PointNet~\cite{qi2017pointnet},  which mainly consists of MLPs and max pooling layers. Considering our point patches are represented by their coordinates, a simple Position Embedding (PE) map the centers $C$ to the embedding $P_c$. Setting the dimension as $d$, the visible tokens $T_v$ is defined as:
\begin{equation}
    T_{vis} = PointNet(P_{vis}), T_{vis} \in \mathbb{R}^{(1-m)n \times d}
\end{equation}
\input{latex/tables/5.3/Algorithm1}

\noindent\textbf{Reconstruction target}.
Our reconstruction backbone is entirely based on Point-Transformer($PT$)~\cite{zhao2021point}, with an asymmetric encoder-decoder architecture. In order to predict the masked points, we add a prediction head with a simple fully connected layer to the last layer of the decoder.

During the training phase, both the visible tokens $T_{vis}$ and the mask tokens $T_m$ are sent to the Transformer $PT$ with the global position embedding $P_c$. At the last layer of the decoder, the prediction head $F_c$ tries to output the reconstructed points $P_{pre}$. The process can be expressed as follows:
\begin{equation}
    P_{pre} = F_c\{PT(T_{vis}, T_m, P_c)\}, P_{pre} \in \mathbb{R}^{mn \times k \times 3}
\end{equation}
The target of our reconstruction net is to restore the masked point patches $M$, also called $P_{gt}$. After obtaining the predicted point patches $P_{pre}$ and the ground truth $P_{gt}$ , we use $l_2$ Chamfer Distance as our reconstruction loss.
\begin{equation}
    L=\frac{1}{\left|P_{p r e}\right|}\sum_{a \in P_{p r e}}\min _{b \in P_{g t}}\|a-b\|_2^2+\frac{1}{\left|P_{g t}\right|}\sum_{b \in P_{g t}}\min _{a \in P_{p r e}}\|a-b\|_2^2
\end{equation}

During testing, each restored point-patches will be concatenated with the visible point-patches and sent back to the Transformer for a few times until the anomaly parts have been masked and restored to normal surface. Given the anomaly samples as $P^*$, the iterative reconstruction pseudo-code is shown in Algorithm 1.

\subsection{Dense feature concatenation and comparision}
Following the iterative reconstruction module, the anomaly score is determined by comparing the reconstructed point cloud with the original input. To enhance detection of subtle anomalies, we combine features from both the input and reconstructed point clouds. Additionally, to reduce false positives caused by excessive iterations in normal point cloud generation, we employ a single template from the training phase to regularize the output.
\label{ssec:ktformer}

\noindent\textbf{Point-Patch comparision.}
Leveraging the point-patch subdivision approach inspired by PatchCore~\cite{roth2022towards}, we address the challenges of the unordered nature of point clouds by facilitating patch-wise comparison between the reconstructed and original point clouds. Each point's anomaly score is derived by transforming the Chamfer distance-based comparison scores of its corresponding patches, computed directly from the training phase. The comparison process is formalized as:
\begin{equation}
    P_i = KNN(p_i,k),  {P_i} \in\mathbb{R}^{n \times k  \times  3}
\end{equation}
\begin{equation}
    P_o = KNN(p_o,k),  {P_o} \in\mathbb{R}^{n \times k  \times  3}
\end{equation}
\begin{equation}
    A_p = \mathcal{L}_\text{c}(P_i,P_o)
\end{equation}
where $p_i$ and $p_o$ denote the input and output point clouds, respectively, with $n$ representing the aggregation count of neighboring points around each point. $P_i$ and $P_o$ are the corresponding point-patches. The Chamfer loss is given by $\mathcal{L}_{\text{c}}$, and $A_p$ is the anomaly score in the point domain.

\noindent\textbf{Feature fusion and comparision.}
It is commonly recognized that features extracted from different layers of a neural network represent information at varying levels. Consequently, the fusion of information from different layers can effectively enhance the capability of anomaly detection. In our DFC (Dense Feature Concatenation and Comparison) module, we utilize the decoder of a transformer previously employed for reconstruction as a feature extractor. Following the decoding of input and reconstructed point clouds, we extract their 1st, 2nd, and 3rd layer features ($f^1,f^2,f^3$) and subsequently fuse and compare them, resulting in feature-level anomaly scores. Feature level anomaly scores $A_f$ are formulated as follows:
\begin{equation}
    f^{1},f^{2},f^{3} = \phi(p)
\end{equation}
\begin{equation}  
    F = f^{1}\oplus f^{2}\oplus f^{3}
\end{equation}
\begin{equation}
    A_f = F_i \Theta F_o
\end{equation}
where $p$ are input and reconstructed point clouds, and $f$ are the extracted features. $\oplus$ represents the fusion operation and $F$ are the fused features. $\Theta$ is the comparision operation and $A_f$ is the feature anomaly score.

\input{latex/tables/5.3/I-AUC}

\noindent\textbf{Template Regularization.}
Excessively iterative reconstruction may induce "normal point drift", potentially increasing the positive false rate. To mitigate this, we employ a feature template $T_f$ saved during the training phase to regularize the features $F_o$ of our reconstructed point cloud. For each vector $z_i$ within $F_o$, we compute its distance to the template's corresponding vector $z_i$ and save the distance to a memory bank $M$.
\begin{equation}
\forall z_i \in \mathcal{F_O},  M = ||\hat{z}^{(l)}_i - z^{(l)}_i||
\end{equation}
By Setting a distance threshold $\tau$, we access each distance $d_i$ within the set $M$. If $d_i$ exceeds $\tau$, we replace the corresponding vector $z_i$ with $\hat z_i$. If using template regularization, the regularized
$F_o$ will be used for calculating $A_f$.

After obtaining the anomaly score $A_p$ and $A_f$, we interpolate them to a uniform dimension. The final anomaly score is the result of concatenating two scores.
\begin{equation}
A = A_p \oplus A_f
\end{equation}
\input{latex/figures/Visulization/As}

%% file: latex/tables/5.3/Algorithm1.tex
\definecolor{dkgreen}{rgb}{0,0.6,0}
\definecolor{gray}{rgb}{0.5,0,0}
\definecolor{mauve}{rgb}{0.58,0,0.82}

\lstset{
	language=Python,
	basicstyle=\small\ttfamily\bfseries,
	numberstyle=\color{gray},
	xleftmargin=5pt,
	framexleftmargin=5pt,
	commentstyle=\color{dkgreen},
	keywordstyle=\color{blue},
}
%

\begin{algorithm}[t]
\caption{Iterative Mask Reconstruction}
\label{algo1}
\begin{algorithmic}[1]
\Require Point Transformer $PT$, Prediction head $F_c$, Geometry feature extractor $G$, Feature related sample function $S$, masking ratio $m$, iteration number $I$, data loader $\mathcal{D}$
\State Load model $PT$
\State Set $PT$ to evaluation mode
\For {each $P^{*}$ in $\mathcal{D}$}
\State $(P, C) \gets \text{GPS}(P^{*})$
\For {index in $1$ to $I$}
\State $P_{\text{vis}} \gets P(1 - m)$
\State $T \gets PT(P_{\text{vis}}, C)$
\State $P_{pre} \gets F_c(T, C)$
\State $P = P_{pre} \oplus P_{vis}$
\EndFor
\EndFor
\Function{GPS}{$P$}
\State $M \gets G(P)$
\State $C \gets S(M, P)$
\State $P \gets \text{KNN}(C, P, k)$
\State \Return $(P, C)$
\EndFunction
\end{algorithmic}
\end{algorithm}

%% file: latex/tables/5.3/I-AUC.tex
\begin{table*}
  \centering
  \scriptsize
  \renewcommand\arraystretch{0.55}
  \begin{tabular}{@{}cl|m{0.6cm}<{\centering}m{0.6cm}<{\centering}m{0.6cm}<{\centering}m{0.6cm}<{\centering}m{0.6cm}<{\centering}m{0.6cm}<{\centering}m{0.6cm}<{\centering}m{0.65cm}<{\centering}m{0.65cm}<{\centering}m{0.6cm}<{\centering}m{0.6cm}<{\centering}m{0.6cm}<{\centering}|m{0.6cm}<{\centering}}
    \toprule
    & Method & Airplane & Car & Candy & Chicken & Diamond & Duck & Fish & Gemstone & Seahorse & Shell & Starfish& Toffees & Mean\\
    \midrule 
     &BTF(Raw) & 0.730 & 0.647 & 0.539 & 0.789 & 0.707 & \bf{0.691} & 0.602 & \bf{0.686} & 0.596 & 0.396 & 0.530&0.703 & 0.635\\
     &BTF(FPFH) & 0.520 & 0.560 & 0.630 & 0.432 & 0.545 & 0.784 & 0.549 & 0.648 & \bf{0.779} & \bf{0.754} & 0.575 &0.462 &0.603 \\
     &M3DM & 0.434 & 0.541 & 0.552 & 0.683 & 0.602 & 0.433 & 0.540 & 0.644 & 0.495 & 0.694 & 0.551&0.450 &0.552 \\
     &PatchCore(FPFH) & \bf{0.882} & 0.590 & 0.541&0.837 & 0.574 & 0.546 & 0.675 & 0.370 & 0.505 & 0.589 & 0.441&0.565 &0.593 \\
     &PatchCore(PointMAE) & 0.726  & 0.498 & 0.663 & 0.827 & 0.783 & 0.489 & 0.630 & 0.374 & 0.539 &0.501 & 0.519&0.585 &0.594 \\
     &CPMF & 0.701 & 0.551 & 0.552 & 0.504 & 0.523 & 0.582 & 0.558 & 0.589 & 0.729 & 0.653 & \bf{0.700}&0.390 &0.586\\
     &RegAD & 0.716 & 0.697 & 0.685 & \bf{0.852} & 0.900 &  0.584 & {\bf 0.915} & 0.417 & 0.762 & 0.583 & 0.506&\bf{0.827} &0.704\\
     &\textbf{Ours} & 0.762 & \bf{0.711} & {\bf 0.755} &  0.780 & {\bf 0.905} & 0.517 & 0.880 & 0.674& 0.604 & 0.665 & 0.674 & 0.774&\bf{0.725}\\
    \bottomrule
  \end{tabular}
  \vspace{-5pt}
  \caption{\textbf{I-AUROC score for anomaly detection of 12 categories of Real3D-AD}. Bold numbers represent the current highest metrics. Our method clearly outperforms the baseline; For pure 3D point setting, we get 0.725 mean I-AUROC score.}
  \vspace{3pt}
  \label{tab:iaucroc1}
\end{table*}

\begin{table*}
  \vspace{-10pt}
  \centering
  \scriptsize
  \renewcommand\arraystretch{0.55}
  \begin{tabular}{@{}cl|m{0.6cm}<{\centering}m{0.6cm}<{\centering}m{0.6cm}<{\centering}m{0.6cm}<{\centering}m{0.6cm}<{\centering}m{0.6cm}<{\centering}m{0.6cm}<{\centering}m{0.65cm}<{\centering}m{0.65cm}<{\centering}m{0.6cm}<{\centering}m{0.6cm}<{\centering}m{0.6cm}<{\centering}m{0.6cm}<{\centering} m{0.6cm}<{\centering}}
    \toprule
    & Method & cap0 & cap3 &helmet3  & cup0 & bowl4 & vase3 & headset1 & eraser0 & vase8 & cap4 &vase2 & vase4 & helmet0 & bucket1\\
    \midrule 
     &BTF(Raw) & 0.668 & 0.527 & 0.526 & 0.403 & 0.664 & \bf{0.717} & 0.515 & 0.525 & 0.424 & 0.468 & 0.410&0.425 &0.553&0.321 \\
     &BTF(FPFH) & 0.618 & 0.522 & 0.444 & 0.586 & 0.609 & 0.699 & 0.490 & \bf{0.719} & \bf{0.668} & 0.520 & 0.546 &0.510 &0.571&0.633 \\
     &M3DM & 0.557 & 0.423 & 0.374 & 0.539 & 0.464 & 0.439 & 0.617 & 0.627 & 0.663 & \bf{0.777} & 0.737&0.476 &0.526&0.501 \\
     &Patchcore(FPFH) & 0.580 & 0.453 & 0.404& 0.600 & 0.494 & 0.449 & 0.637 & 0.657 & 0.662 & 0.757 & 0.721&0.506 &0.546&0.551 \\
     &Patchcore(PointMAE) & 0.589  & 0.476 & 0.424 & 0.610 & 0.501 & 0.460 & 0.627 & 0.677 & 0.663 &0.727 & \bf{0.741}&0.516 &0.556&0.561 \\
     &CPMF & 0.601 & 0.551 & 0.520 & 0.497 & \bf{0.683} & 0.582 & 0.458 & 0.689 & 0.529 & 0.553 & 0.582&0.514 &0.555&0.601\\
     &RegAD & 0.693 & 0.725 & 0.367 &0.510& 0.663 & 0.650 & 0.610 & 0.343 & 0.620 & 0.643 & 0.605 & 0.500&\bf{0.600} &0.752\\
     &\textbf{Ours} & {\bf 0.737} & \bf{0.775} & {\bf 0.573} & {\bf 0.643} & 0.676 & 0.700  & 0.676 & 0.548 &0.630 & 0.652 &0.614 &\bf{0.524} & 0.597&\bf{0.771}\\
     \midrule
    \midrule
    & Method & bottle3 & vase0 & bottle0 & tap1 & bowl0 & bucket0 & vase5 & vase1 & vase9 & ashtray0 & bottle1 & tap0 & phone & cup1\\
    \midrule 
     &BTF(Raw) & 0.568 & 0.531 & 0.597 & 0.573 & 0.564 & \bf{0.617} & 0.585 & 0.549 & 0.564 & 0.578 & 0.510&0.525 &0.563&0.521 \\
     &BTF(FPFH) & 0.322 & 0.342 & 0.344 & 0.546 & 0.509 & 0.401 & 0.409 & 0.219 & 0.268 & 0.420 & 0.546 &0.560 &0.671&0.610 \\
     &M3DM & 0.541 & 0.423 & 0.574 & 0.739 & 0.634 & 0.309 & 0.317 & 0.427 & \bf{0.663} & 0.577 & 0.637&\bf{0.754} &0.357&0.556 \\
     &Patchcore(FPFH) & 0.572 & 0.455 & \bf{0.604}&  \bf{0.766} & 0.504 & 0.469 & 0.417 & 0.423 & 0.660 & 0.587 & 0.667&0.753 &0.388&0.586 \\
     &Patchcore(PointMAE) & \bf{0.650}  & 0.447 & 0.513 & 0.538 & 0.523 & 0.593 & 0.579 & 0.552 & 0.629 &0.591 & 0.601&0.458 &0.488&0.556 \\
     &CPMF & 0.405 & 0.451 & 0.520 & 0.697 & \bf{0.783} & 0.482 & 0.618 & 0.345 & 0.609 & 0.353 & 0.482&0.359 &0.509&0.499\\
     &RegAD & 0.525 & 0.533 & 0.486 & 0.641 & 0.671 & 0.610 & 0.520 & 0.702 & 0.594 & 0.597 & 0.695&0.676 &0.414&0.538\\
     &\textbf{Ours} & 0.640 & \bf{0.533} & {\bf 0.552} & 0.696 &  0.681 & 0.580 & \bf{0.676} & \bf{0.757}&0.594&\bf{0.671}& \bf{0.700} & 0.676&\bf{0.755} &\bf{0.757} \\
    \midrule
    \midrule
    & Method & vase7 & helmet2  & cap5 & shelf0 & bowl5 & bowl3 & helmet1 & bowl1 & headset0 & bag0& bowl2 & jar& \multicolumn{2}{|c}{\textbf{Mean}} \\
    \midrule 
     &BTF(Raw) & 0.448 & 0.602  & 0.373 & 0.164 & 0.417 & 0.385 & 0.349 & 0.264 & 0.378 & 0.410&0.525 &0.420&\multicolumn{2}{|c}{0.493} \\
     &BTF(FPFH) & 0.518 & 0.542  & 0.586 & 0.609 & 0.699 & 0.490 & \bf{0.719} & 0.668 & 0.520 & 0.546 &0.510 &0.424&\multicolumn{2}{|c}{0.528} \\
     &M3DM & 0.657 & 0.623  & 0.639 & 0.564 & 0.409 & 0.617 & 0.427 & 0.663 & 0.577 & 0.537&0.684 &0.441&\multicolumn{2}{|c}{0.552} \\
     &Patchcore(FPFH) & \bf{0.693} & 0.425 &  \bf{0.790} & 0.494 & 0.558 & 0.537 & 0.484 & 0.639 & 0.583 & 0.571&0.615 &0.472&\multicolumn{2}{|c}{0.568} \\
     &Patchcore(PointMAE) & 0.650  & 0.447  & 0.538 & 0.523 & 0.593 & 0.579 & 0.552 & 0.629 &0.591 & 0.601&0.458 &0.483&\multicolumn{2}{|c}{0.562} \\
     &CPMF & 0.397 & 0.462  & 0.697 & 0.685 & 0.685 & \bf{0.658} & 0.589 & 0.639 & 0.643 & 0.643&0.625 &0.610&\multicolumn{2}{|c}{0.559}\\
     &RegAD & 0.462& 0.614  & 0.467 &\bf{0.688} & 0.593 & 0.348 & 0.381 & 0.525 & 0.537 & \bf{0.706}&0.490 &0.592&\multicolumn{2}{|c}{0.572}\\
     &\textbf{Ours} & 0.635 & \bf{0.641}  & 0.652 &  0.603 & \bf{0.710} & 0.599 & 0.600& \bf{0.702} & {\bf 0.720} &0.660 &\bf{0.685} &\bf{0.780} & \multicolumn{2}{|c}{\bf{0.661}}\\
    \bottomrule
  \end{tabular}
  \vspace{-5pt}
  \caption{\textbf{I-AUROC score for anomaly detection of 40 categories of our Anomaly-ShapeNet dataset}. Our method clearly outperforms other methods. Last line is the average result of 40 classes. The results can be regarded as the baseline of Anomaly-ShapeNet.}
  \vspace{-10pt}
  \label{tab:iaucroc2}
\end{table*}

%% file: latex/figures/Visulization/As.tex
\begin{figure}
\begin{center}
\hspace{-3mm}
\subfloat{\includegraphics[width=\linewidth]{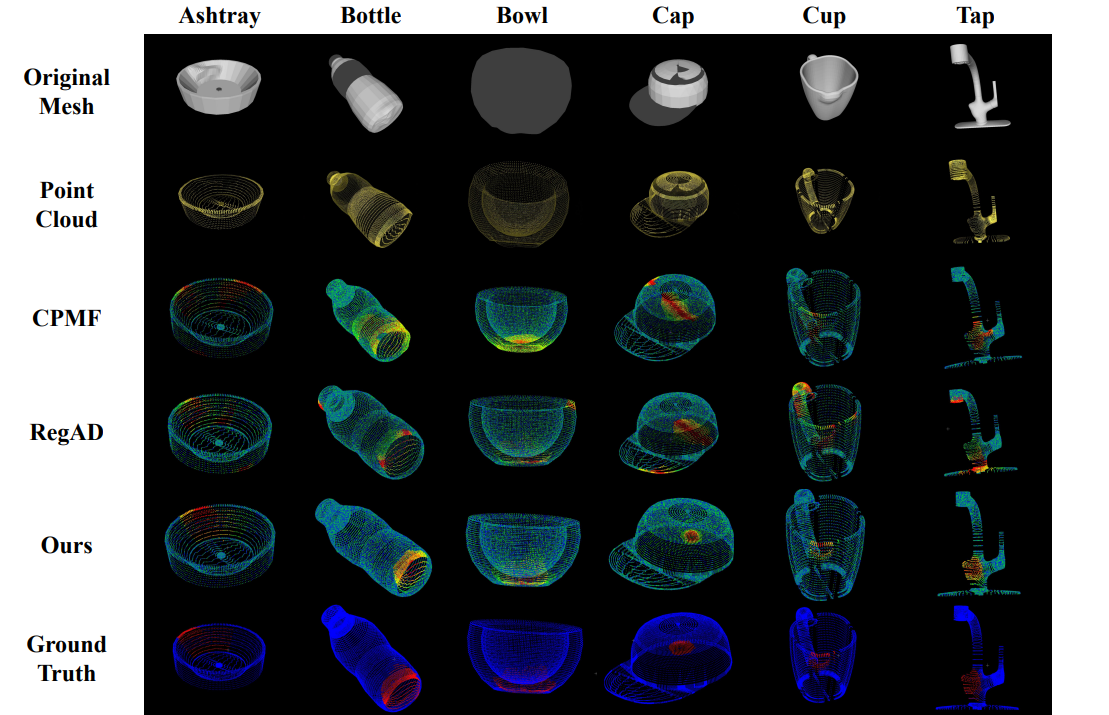}}\end{center}
\vspace{-4mm}
\caption{\textbf{Qualitative results visualization of anomaly localization performance on Anomaly-ShapeNet.}}
\vspace{-5mm}
\label{fig:Vis0}
\end{figure}


%% file: latex/sections/5_experiments.tex
\section{Experiments}
\subsection{Datasets}
\noindent\textbf{Anomaly-ShapeNet.}
The Anomaly-ShapeNet is our newly proposed 3D synthesised point cloud anomaly detection dataset. The Anomaly-ShapeNet dataset offers 40 categories, with more than 1600 positive and negative samples. Each training set for a category contains only four samples, which is similar to the few-shot scenario. Each test set for a category contains normal and various defected samples. Meanwhile, we evaluated M3DM~\cite{wang2023multimodal}, RegAD~\cite{huang2022registration}, and other methods on our designed dataset, providing a benchmark for future researchers.\\
\noindent\textbf{Real3D-AD.}
Real3D-AD~\cite{wang2023multimodal} is a new large-scale 3D point anomaly dataset captured with the PMAX-S130 high-resolution binocular 3D scanner. It contains 1,254 samples across 12 categories and includes the Reg-AD as a baseline.

\subsection{Evaluation metrics}
Image-level anomaly detection is measured using I-AUROC (Area Under the Receiver Operator Curve), with higher values indicating superior detection capabilities. Pixel-level anomalies are evaluated via the same curve for segmentation accuracy. Besides, the per-region overlap (AUPRO) metric and testing time results are also provided in the Appendix.
\subsection{Implementation details}
The backbone architecture used in our experiments is directly adopted from Point-MAE~\cite{pang2022masked}. Instead of training directly on our dataset, we first train the backbone using ShapeNet-55~\cite{chang2015shapenet} with a point size of 8192. Regarding the geometry-aware sample module, we set the threshold $\tau$ to 0.3, and the number of points sampled in salient regions is twice that of non-salient regions. When fine-tuning our model on both the Anomaly-ShapeNet and Real3D-AD datasets, we convert the point clouds into a $256 \times 64$ point-patch structure. Here, 256 represents the number of central points sampled, and 64 represents the number of neighborhood points selected using K-nearest neighbors (KNN). During training and testing, we set the mask rate to 0.4, and the number of iterations for testing is 3. For anomaly scoring, we utilize the $1st$, $2nd$, and $3rd$ intermediate layers of the backbone's decoder for comparison. 
\subsection{Anomaly detection on Anomaly-ShapeNet and Real3D-AD Datasets}
Anomaly detection results on Anomaly-ShapeNet and Real3D-AD are shown in table~\ref{tab:iaucroc1} and table~\ref{tab:iaucroc2}. Compared to other methods, our IMR-Net performs better on the average I-AUROC, which achieves 72.5\% on Real3D-AD and 66.1\% on Anomaly ShapeNet. In table~\ref{tab:iaucroc2}, our IMR-Net achieves the highest score for 19 out of 40 classes. We visualize some representative samples of Anomaly-ShapeNet for anomaly detection and localization in  Figure~\ref{fig:Vis0} . 

\subsection{Ablation study}

\noindent\textbf{Effectiveness of geometry aware sampling.}
FPS (farthest point sampling) and RS (random samping) are widely used in 3D point processing. 
However, when the sampling ratio is too low, FPS (Farthest Point Sampling) and RS (Random Sampling) may result in the sampled defects too slightly to be detect. Therefore, we employ the geometry aware sampling. To demonstrate the superiority of the GPS algorithm in anomaly detection, we conducted ablation experiments alongside RS , FPS , and Voxel Down-sampling methods. As shown in table \ref{table:downsample} , when using our GPS, the higher I-AUC (0.66) and P-AUC (0.65) were achieved.

\noindent\textbf{Analysis of masking ratio.}
Figure~\ref{fig:mask} shows the influence of masking ratio. The optimal ratio is 0.4, which is good both for reconstruction and feature representation. When the masking rate is reduced, the anomaly regions may not be covered during the iteration process. Conversely, when the masking ratio increases, the limited data may prevent the model from convergence. Figure~\ref{fig:mask1} analyzes the relationship between the size of the anomalous region and the masking rate, where we find that point clouds with larger anomalous areas require higher masking rates.

\noindent\textbf{Feature Discrimination ability.}
Previous methods like M3DM~\cite{wang2023multimodal} and RegAD~\cite{huang2022registration} primarily employed pre-trained models on other datasets to extract features, leading to domain bias in the extracted features compared to the actual anomaly detection datasets. In contrast, our IMRNet, as a self-supervised network, effectively extracts features from both the abnormal and normal point clouds in the Real3D-AD and Anomaly-ShapeNet datasets. This is directly proved in Figure~\ref{fig:duck}. Our extracted features have a more compacted feature space and this property makes the features more suitable for memory bank building and feature distance calculation.


\input{latex/tables/5.3/Downsample}
\input{latex/figures/AS/iteration}
\input{latex/figures/Visulization/Tsne}

%% file: latex/tables/5.3/Downsample.tex
\begin{table}[!t]
\small
\centering
  \renewcommand\arraystretch{0.7}
\setlength\tabcolsep{1.0pt}
\def\arraystretch{0.7}
\resizebox{0.7\linewidth}{!}{
\begin{tabular}{L{2.0cm}|C{1.1cm}|C{1.1cm}}
\hline
\,~\multirow{2}{*}{\# Sample } &  \multicolumn{2}{c}{Metrics} \\
\cline{2-3}
& I-AUC & P-AUC    \\
\hline
\;~RS & 0.55 & 0.61    \\
\;~FPS & 0.64 & 0.62    \\
\;~Voxel & 0.62 & 0.55    \\
\;~\textbf{GPS} & \bf{0.66} & \bf{0.65}    \\
\hline
\end{tabular}}
\caption{
    \textbf{Ablation study on the sample methods.}
    The bold number represents the sample method corresponding to the highest index. RS represents random sampling and FPS represents farthest random sampling. Voxel denotes voxel down-sample and GPS is our geometry aware sampling. 
}
\vspace{-3mm}
\label{table:downsample}
\end{table}

%% file: latex/figures/AS/iteration.tex



\begin{figure}
    \subfloat[Overall Performance.]{
        \begin{minipage}[t]{0.46\linewidth}
            \centering
            \includegraphics[width=0.9\linewidth]{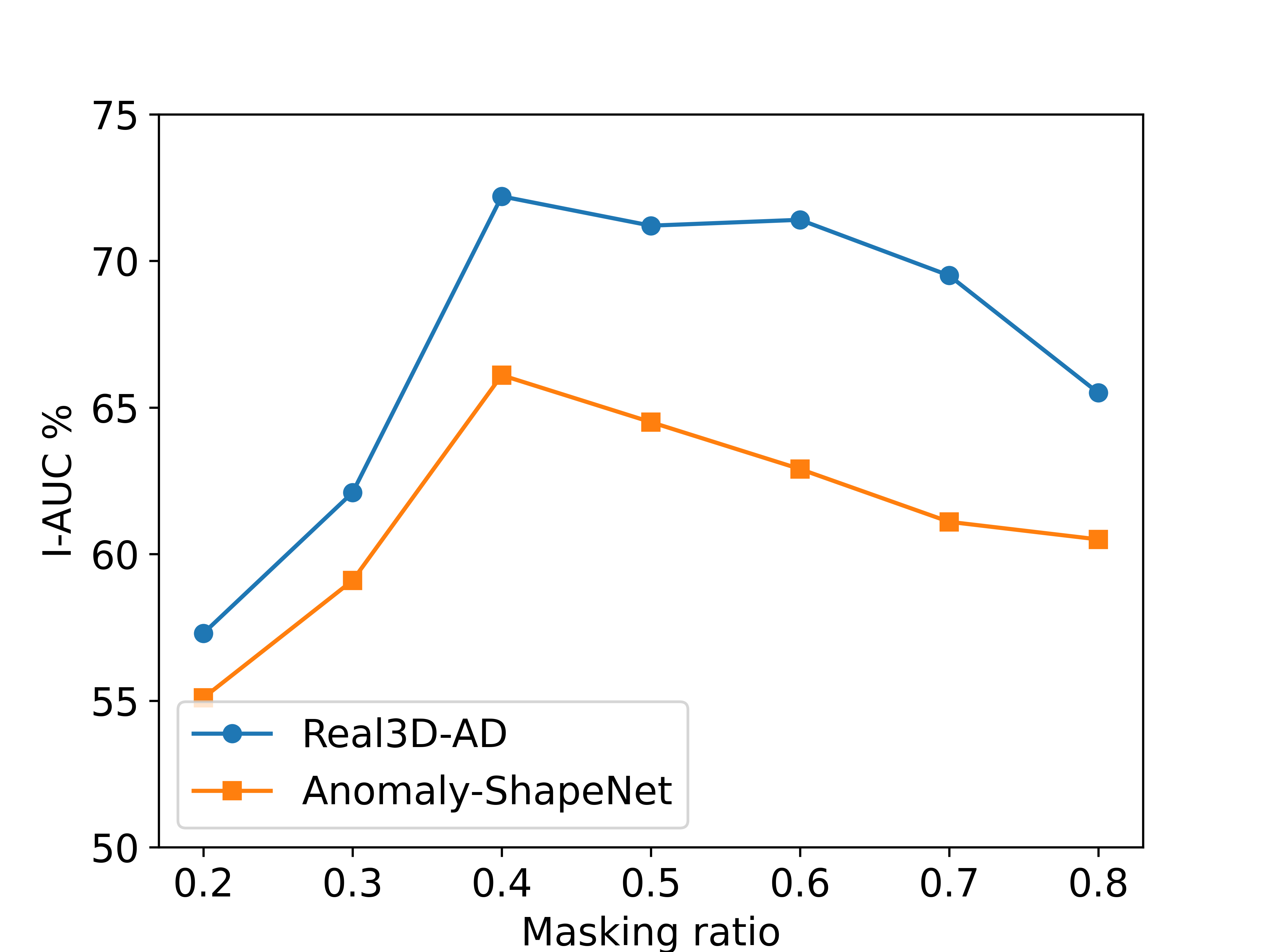}
            \label{fig:mask0}
        \end{minipage}
    }
    \hfill
    \subfloat[Categories Performance.]{
        \begin{minipage}[t]{0.46\linewidth}
            \centering
            \includegraphics[width=0.9\linewidth]{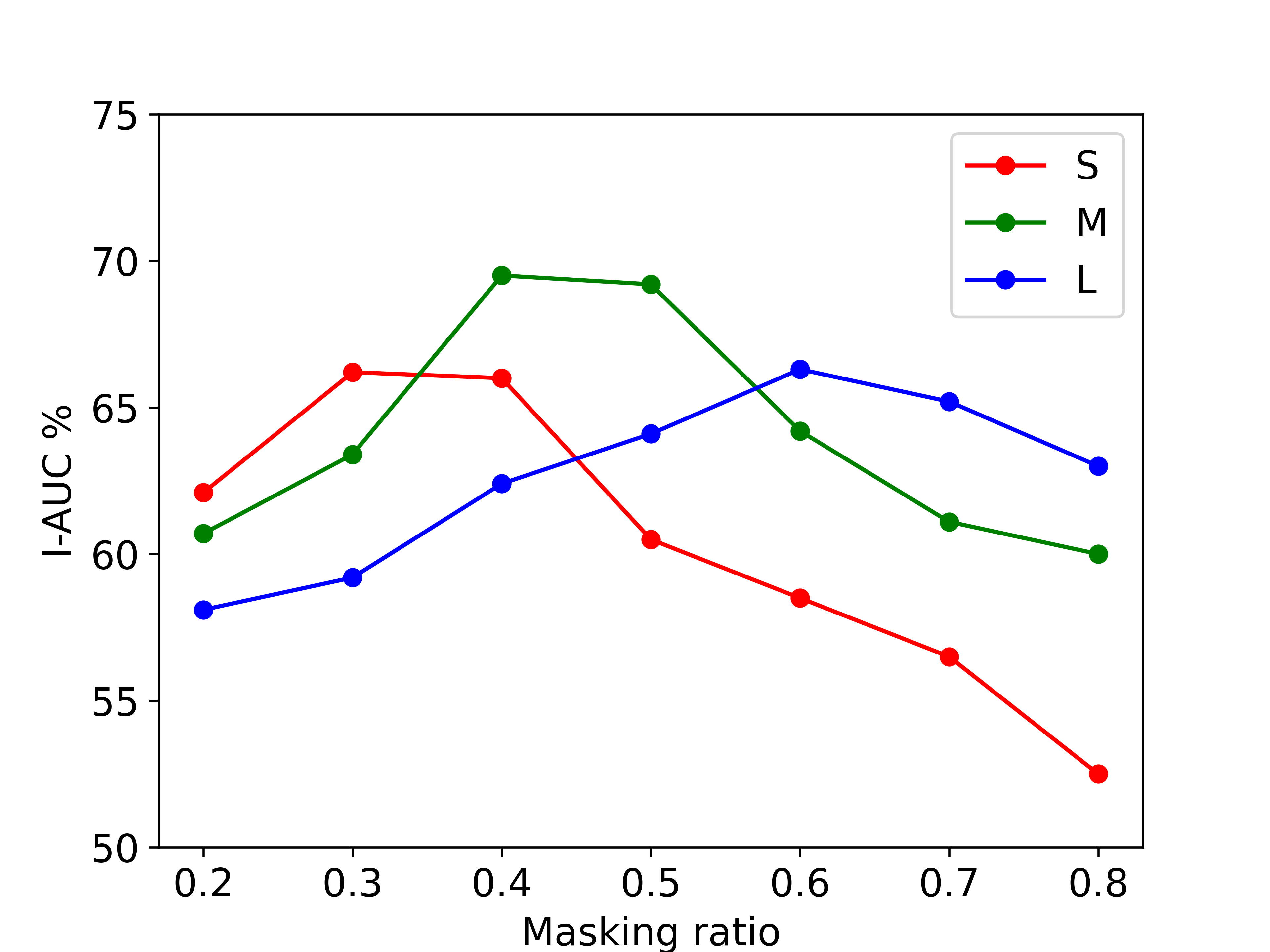}
            \label{fig:mask1}
        \end{minipage}
    }
    \vspace{-5pt}
     \caption{Ablation study of masking ratio. (S,M,L) represent (small,middle,large) size anomaly. At the masking ratio 0.4, the overall performance is best. Objects with larger anomaly correspond to higher optimal masking ratios. }
      \vspace{-5pt}
     \label{fig:mask}
\end{figure}

%% file: latex/figures/Visulization/Tsne.tex
\begin{figure}\centering
\vspace{-3mm}
\subfloat{\includegraphics[width=0.8\linewidth]{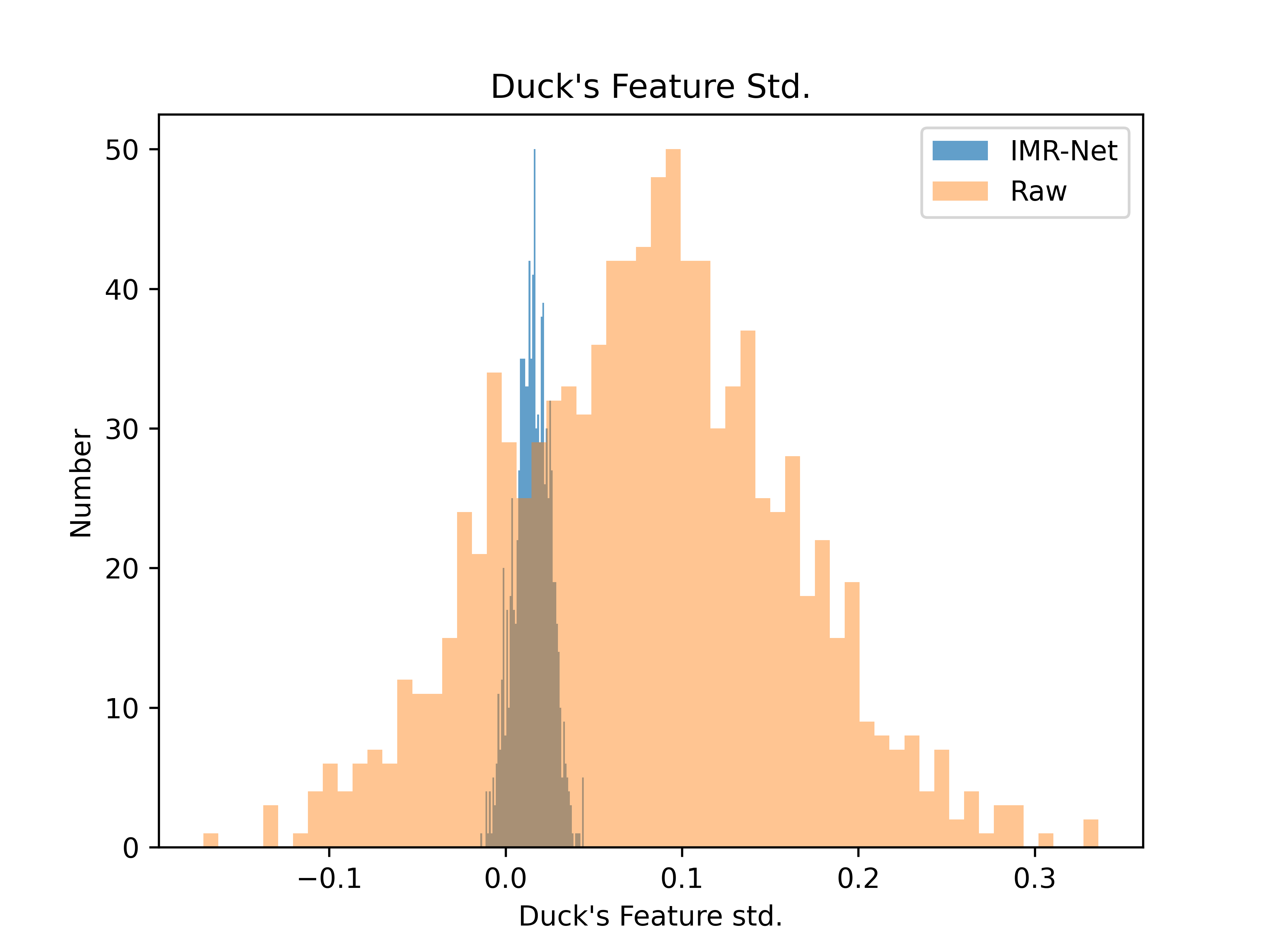}}
\caption{Histogram of standard deviation along each dimension of feature of extracted by pretrained model and our IMRNet model. We show a case of the \textit{Duck} class in Real3D-AD dataset.}
\label{fig:duck}
%

%

\vspace{-5mm}

\end{figure}

%% file: latex/sections/6_conclusion.tex
\section{Conclusion}
In this work, we propose Anomaly-ShapeNet, a synthetic 3D point dataset for anomaly detection, containing realistic and challenging samples. The diverse point clouds with high accuracy and reasonable quantity in Anomaly-ShapeNet make it more suitable for various 3D algorithms. Moreover, we firstly introduce IMRNet, a self-supervised model based on 3D point mask reconstruction, achieving the state-of-the-art performance on both the Anomaly-ShapeNet and Real3D-AD datasets.
